\documentclass{esannV2}
\usepackage{graphicx}
\usepackage[latin1]{inputenc}
\usepackage{amssymb,amsmath,array}

\usepackage{fancyhdr}
\fancypagestyle{plain}{%
   \fancyhf{} 
   \fancyfoot[C]{\scriptsize As published in the proceedings of the $27^{th}$ European Symposium on Artificial Neural Networks,\\Computational Intelligence and Machine Learning (ESANN), pages 179-184, 2019.}
   
}

\begin{document}


\title{Fast and Reliable Architecture Selection for Convolutional Neural Networks}

\author{Lukas Hahn$^{1, 2}$, Lutz Roese-Koerner$^1$, Klaus Friedrichs$^1$ and Anton Kummert$^2$
%
%
\vspace{.3cm}\\
%
1- Aptiv \\
Wuppertal, Germany
%
\vspace{.1cm}\\
2- University of Wuppertal - Department of Electrical Engineering \\
Wuppertal, Germany\\
}

\maketitle
\vfill

\begin{abstract}
The performance of a Convolutional Neural Network (CNN) depends on its hyperparameters, like the number of layers, kernel sizes, or the learning rate for example. Especially in smaller networks and applications with limited computational resources, optimisation is key.\\ We present a fast and efficient approach for CNN architecture selection. Taking into account time consumption, precision and robustness, we develop a heuristic to quickly and reliably assess a network's performance. In combination with Bayesian optimisation (BO), to effectively cover the vast parameter space, our contribution offers a plain and powerful architecture search for this machine learning technique.
\end{abstract}

\section{Introduction}
\label{sec:intro}

Finding the optimal architecture for a Convolutional Neural Network (CNN) has been a challenging task since the introduction of this machine learning technique. The search for improvement of their performance has led to numerous optimisation approaches, with the state of the art constantly and rapidly evolving. 
While the steady increase of available computational capacity in some respects enables researchers to neglect this problem and solve complex learning tasks with extremely deep networks, individual optimisation remains important.
Especially with the application of CNNs to consumer products in various industrial sectors, the need for small, specialised and highly optimised networks which are able to run on  mobile embedded hardware re-emerges.

Considering this challenge it stands to reason, that one would like to develop a method to find the best network architecture for a given task. This method should be fast enough to clearly outperform manual approaches in a matter of time consumption. The faster decision support is provided, the more time remains for fine tuning the outcome, since it is unlikely that final optimisation through human supervision will become obsolete. Two aspects are particularly important to such an architecture search: First a technique to estimate the performance of a candidate architecture is required. Normal training is very time consuming and therefore not recommended. This estimation should not only be fast, but reliable in predicting the outcome which a normal training would have. The second aspect regards the search algorithm itself. Even shallow networks comprise hundreds up to thousands of hyper-parameters which renders the use of an efficient way to cover the vast search space indispensable.

\section{Related Work}
\label{sec:related}
Architecture search and hyperparameter optimisation are two important topics in the field of machine learning, leading to numerous publications. Much work focuses on Bayesian optimisation (BO) \cite{snoek} \cite{bayesian} \cite{mbo}, to address general tasks in selecting hyperparameters. Other authors directly examine the problem of CNN architecture search. We offer our contribution to the latter and combine it with the former to obtain a comprehensive solution. While some papers introduce architecture search methods which are carefully engineered, but often difficult to use in practice \cite{EPANN} \cite{Darts} \cite{ParameterSharing} \cite{esann}, we aim for a plain and effective method to provide grounds for performance estimation and selection decision, that could be applied to nearly any type of neural network. 

Building on the work of \cite{random_jerrett}, Saxe et al. \cite{randomweights} used random weights to perform tests on the influence of different architectures on classification performance. They employed convolutional and non-convolutional unsupervised feature learning and used a linear SVM for classification. 


\section{Methods}

\label{sec:methods}
We propose a method for finding the best CNN architecture for a certain task. This contribution is split into developing a heuristic method estimating the performance of a candidate architecture without the need of full training and applying BO to narrow the search space. 

To achieve fast performance estimation of a candidate architecture, the proposed approach divides the network into two parts. One comprising the convolution layers used to extract features from the input data and the other consisting of the fully connected classification part of the network (cf. Fig. 1). Since the feature extraction part is more computationally expensive in the vast majority of applications, the number of epochs optimising the weights of the convolutions is restricted to two. These epochs are also the ones which provide the greatest improvement in minimising the classification error. Training of the fully connected layers is much less time consuming and can therefore be conducted for a greater number of epochs (e.g. 30), the exact number depending on the given task. To make the procedure robust and to provide an accurate assessment of the performance of an architecture, every candidate is tested five times in that way, with different initialisations of the network's weights.
\begin{figure}[h!]
	\centering
    \includegraphics[keepaspectratio = true, width = \textwidth]{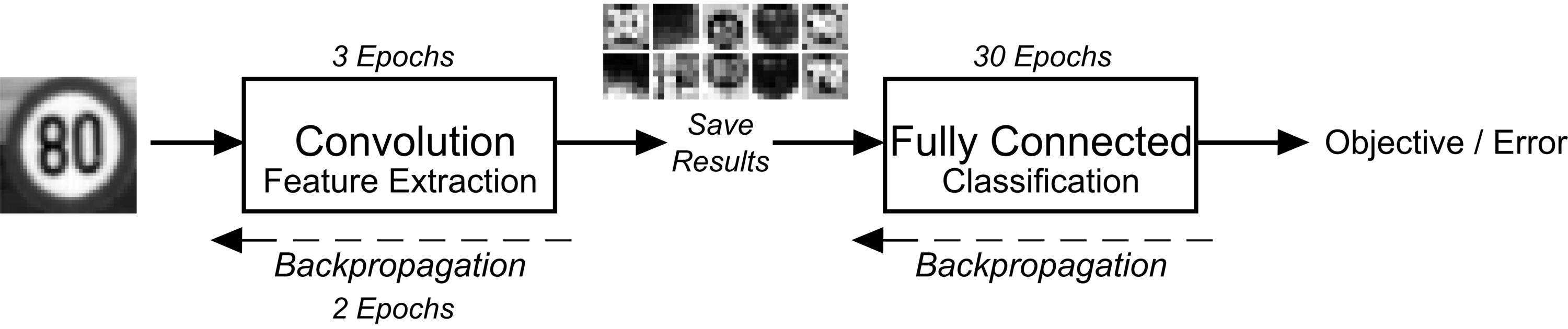}
	\caption{Block diagram describing the proposed heuristic.}
	\label{fig:heuristic}
\end{figure}\\
While the proposed heuristic enables a fast and reliable performance estimation of one candidate architecture, an efficient search method is still required to handle the extensive parameter space. Since exhaustive approaches like grid search are deemed impractical, as the number of combinations to evaluate grows exponentially with the number of hyperparameters and the range of values they can take, the use of a more sophisticated method is advisable. In general, even random search was proven to not only outperform grid search in terms of time consumption, but also provide better results \cite{Bergstra}.

We apply BO, a form of model-based optimisation, as described in \cite{bayesian} and \cite{mbo}. Its combination of random search sampling and improvement driven optimisation facilitates a well-balanced trade-off between exploration and exploitation of the search space and enables a very efficient architecture selection for CNNs. We choose Latin hypercube sampling \cite{latinhypercube} 
as the random sampling method to generate an initial design of the surrogate function based on Kriging models, also known as Gaussian process regression. As infill criterion we select expected improvement, which focuses the iterative selection on promising points, taking into account not only the expected value, but the uncertainty.

\section{Experiments and Results}
\label{sec:experiments}

To validate our claims, we conduct performance estimation and timing experiments with the proposed heuristic and the combined architecture selection. 

\subsection{The Proposed Heuristic}

To determine how well our proposed heuristic estimates the performance of a CNN architecture, we conduct a series of experiments on a number of different datasets for image classification. We select the well-known MNIST \cite{MNIST} and USPS~\cite{USPS} datasets, which provide samples of handwritten digits $0$ to $9$. Further the Latin part of the CoMNIST dataset \cite{CoMNIST} with handwritten letters A to Z. The $278\times278$ pixel source images were resized to $32\times32$ using bicubic interpolation. Finally we use two of our private datasets; a set for traffic sign recognition (TSR), comprising of $40\times40$ images of 19 different traffic signs and a class of objects that are not signs, and one for adaptive headlight control (AHC), with small $13\times13$ image patches of three classes, headlights, taillights and other light sources. As additional input, the AHC dataset uses a vector of $77$ hand crafted features for each sample. Both private datasets provide images with grey and red channel. Details on the datasets can be found in Table \ref{tab:datasets}.

\begin{table}[h!]
	\centering
	\begin{tabular}{l|ccccc}
		& MNIST & CoMNIST & USPS  & TSR    & AHC    \\ \hline \hline
		Classes            & $10$    & $26$      & $10$    & $20$     & $3$      \\
		Image Size         & $28\times28$ & $32\times32$   & $16\times16$ & $40\times40$  & $13\times13$  \\
		Training Samples   & $60\,000$ & $9\,918$    & $10\,000$ & $265\,774$ & $92\,159$  \\
		Validation Samples & $10\,000$ & $1\,300$    & $1\,000$  & $66\,443$  & $115\,477$ \\ \hline
		No. of Architectures & $64$ & $32$ & $96$ & $40$ & $32$ \\
		Corr. Mean-Best & $0.785$ & $0.679$ & $0.828$ & $0.852$ & $0.875$ \\
		Corr. Best-Best & $0.789$ & $0.597$ & $0.742$ & $0.843$ & $0.878$ \\
	\end{tabular}
	\caption{Datasets used for the experiments, including the resulting correlations between the results obtained with our heuristic and the mean value (Corr. Mean-Best) as well as the best value (Corr. Best-Best) of the normal training results.}
	\label{tab:datasets}
\end{table}

We define a number of individual candidate CNN architectures for each dataset. For these experiments we use a strictly deterministic selection of candidate architectures, to assure reproducible and comparable results. Changeable aspects of the structures are for example the number and kernel size of convolutions used, the number of channels per convolutional layer or neurons per hidden layer, the number of fully-connected layers, the type of non-linearity applied between layers, as well as global hyperparameters such as learning rate and momentum.\\
In each case we train every architecture normally ten times, using different random number generator seeds to initialise the weights and biases of the network, to obtain a baseline by taking the mean validation error out of those ten runs. The best value is also considered. For our proposed heuristic we train the feature extraction part of the network for two and the fully-connected part for $15$ or $30$ epochs, depending on whether we selected $50$ or $100$ epochs for the corresponding baseline training. We repeat this five times for every architecture and compare the results by calculating the correlation coefficient between the baseline and the best validation error obtained by our heuristic (cf. Tab. \ref{tab:datasets}.)

Overall, we find a significant correlation between the performance estimation of our heuristic approach and the baseline. The correlation by trend seems to be a little bit higher, if the baseline is considered as the mean out of the normal training runs. This should also be a better representation of the normal use case, where one would train a candidate architecture not ten times, but once.

\begin{figure}[h!]
	\centering
	\includegraphics[keepaspectratio = true, width = 1.0\textwidth]{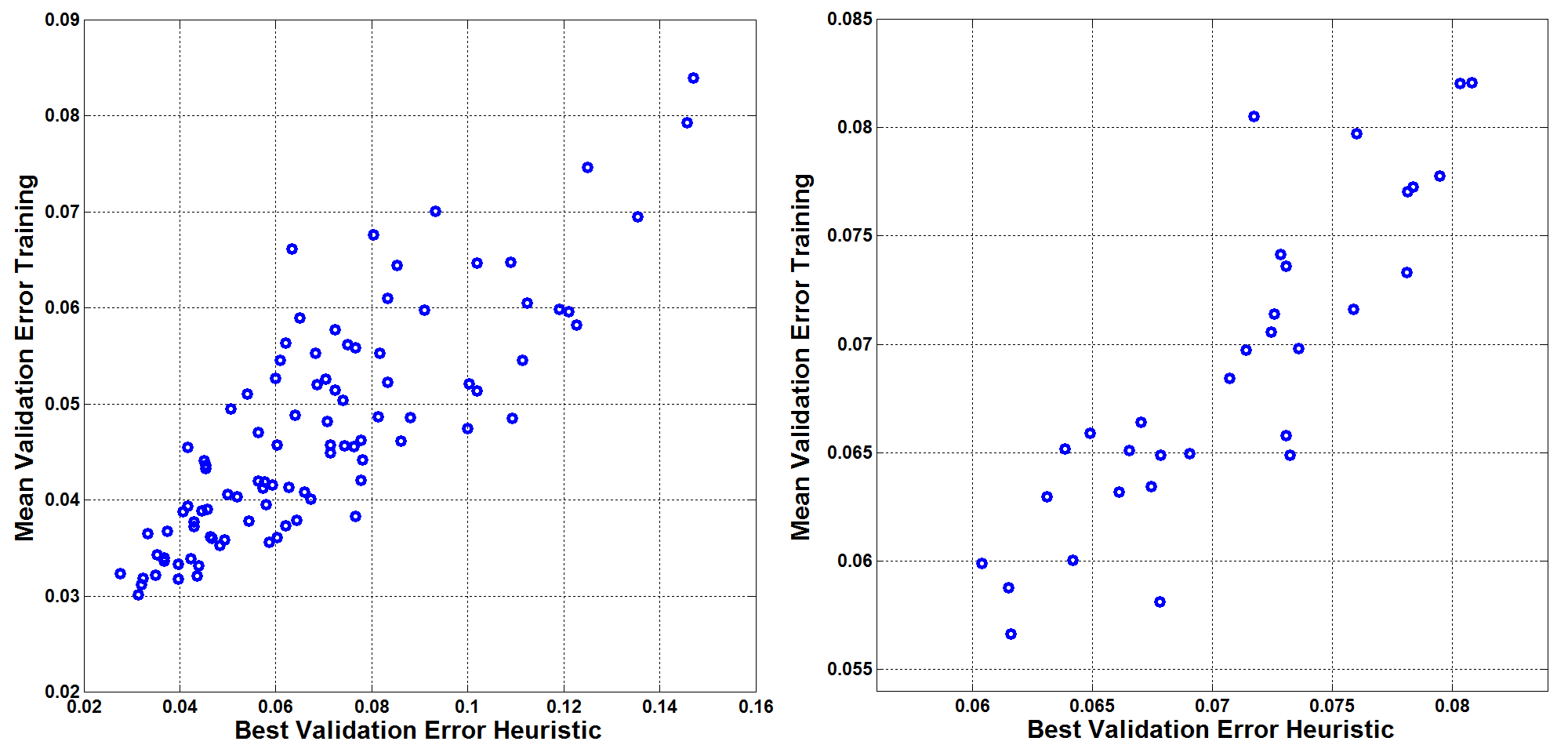}
	\caption{Graph of the experimental correlation of our proposed heuristic and normal training on MNIST (left) and AHC (right).}
	\label{fig:correlation}
\end{figure}

Figure \ref{fig:correlation} shows an exemplary result and correlation graph for the $64$ architectures tested on MNIST. We would like to point out, how close the validation results of our heuristic came to the validation error of the baseline qualitatively. With accuracies close to only one percentage point below full training, this raises the question of how much continuous training of all layers adds to the overall performance of a CNN.

\subsection{Time Measurements}

While prediction accuracy is key to any heuristic approach, the time requirement is the second most important aspect. The proposed heuristic was therefore compared with normal training and the random weights approach of Saxe et al.~\cite{randomweights}. Since we are aware, that the runtime of our heuristic depends on the relative size of the convolutional and fully-connected part of a CNN, the arithmetic mean of time measurements from all architectures in our experiments on the TSR dataset were taken into account. Compared to the normal training baseline, on average five evaluations of our heuristic only take $22.6\%$ of the time needed for a single normal training of the same architecture. \\ The random weights approach takes $89.3\%$ of this time, if ten initialisations per architecture are considered, or $44.7\%$ for five. Both of these numbers were used in \cite{randomweights}, with no preference stated. When we evaluate the time consumption on TSR, we found, that even ten random weights initialisations could not reach the performance of five initialisations of our heuristic (Corr. Mean-Best $0.769$, Corr. Best-Best $0.750$). We further prove, that we are nearly five times as fast as a single full training, while providing a basis for performance assessment, which is much more prone to bad initialisation.

\subsection{Architecture Selection}

We combine our proposed heuristic with BO to utilise the synergy for fast and reliable architecture search. We perform a large-scale experiment with the TSR dataset and compare the use of our heuristic against normal training. Even if we assume generous stepping for numeric parameters, a comparable grid search would have approximately $134.9$ million combinations to compute. For the baseline with normal training, we select $n_{init}=50$ random sampling points for the initial design and $n_{iter}=100$ iterations for the optimisation. Our heuristic only takes a fifth to a fourth of the time, hence we choose $n_{init}=100$ and $n_{iter}=200$, which provides broader parameter exploration while still being significantly faster. Computation of the experiment on CPU took $4708$ minutes ($\approx78$ hours) for the baseline and $2586$~minutes ($\approx43$~hours) with our heuristic. In both cases we took the best parameter configuration obtained and compared them with five full trainings each on TSR. We received a best validation error of $0.85\%$ and a mean validation error of $0.90\%$ for the baseline configuration and respectively $0.63\%$ and $0.84\%$ for the architecture found with our heuristic. 

\section{Conclusion}
\label{sec:conclusion}

We have presented a technique to facilitate fast architecture selection for CNNs. It comprises of a heuristic enabling fast, accurate and reliable performance estimation of a candidate architecture, paired with the ability of BO to concentrate the search on promising regions in the parameter space. This combination allows for strong synergy. Through multiple initialisations of a single candidate architecture the heuristic provides BO with results, which are more robust and credible than the outcome of one single training. Thanks to this fact, it was shown that BO combined with the heuristic was not only able to achieve results comparable to those reached by BO and normal training, but even outperformed this slower approach.


\begin{footnotesize}





\begin{thebibliography}{99}


\bibitem{snoek} J. Snoek, H. Larochelle and R. P. Adams, Practical Bayesian Optimization of Machine Learning Algorithms. In Advances in Neural Information Processing Systems 25 (NIPS), pages 2951-2959, 2012. 

\bibitem{bayesian} B. Shahriari, K. Swersky, Z. Wang, R. P. Adams and N. de Freitas, Taking the Human Out of the Loop: A Review of Bayesian Optimization. In proceedings of the IEEE, 104(1):148-175, 2016


\bibitem{mbo} C. Weihs, S. Herbrandt, N. Bauer, K. Friedrichs, and D. Horn. Efficient global optimization: Motivation, variations, and applications. \emph{Archives of Data Science, Series A (Online First)}, 2:1-26, 2017.

\bibitem{EPANN} J.-M. Pérez-Rùa, M. Bacchouche and S. Pateux, Efficient Progressive Neural Architecture Search. In proceedings of the $29^{th}$ British Machine Vision Conference (BMVC), 2018

\bibitem{Darts} H. Liu, K. Simonyan and Y. Yang, DARTS: Differentiable Architecture Search. \emph{arXiv preprint arXiv:1806.09055}, 2018

\bibitem{ParameterSharing} H. Pham, M. Guan, B. Zoph and Q. Le, Efficient Neural Architecture Search via Parameters Sharing. In proceedings of the $35^{th}$ International Conference on Machine Learning (ICML), pages 4095-4104, 2018

\bibitem{esann} D. Ramachandram, M. Lisicki, T. Shields, M. Amer, G. Taylor, Structure optimization for deep multimodal fusion networks using graph-induced kernels. In proceedings of the $25^{th}$ European Symposium on Artificial Neural Networks (ESANN), pages 11-16, 2018

\bibitem{random_jerrett} K. Jarrett, K. Kavukcuoglu, M. Ranzato, and Y. LeCun, What is the best multistage architecture for object recognition? In proceedings of the IEEE $12^{th}$ International Conference on Computer Vision (ICCV), pages 2146-2153, 2009

\bibitem{randomweights} A. Saxe, P. Koh, Z. Chen, M. Bhand, B. Suresh, and A. Ng. On random weights and unsupervised feature learning. In proceedings of the $28^{th}$ international conference on machine learning (ICML), pages 1089-1096, 2011


\bibitem{Bergstra} J. Bergstra and Y. Bengio, Random Search for Hyper-Parameter Optimization. \emph{Journal of Machine Learning Research}, 13:281-305, 2012

\bibitem{latinhypercube} M. Mckay, R. Beckman, and W. Conover, A comparison of three methods for selecting values of input variables in the analysis of output from a computer code. \emph{Technometrics}, 42:55-61, 1979.




\bibitem{MNIST} Y. LeCun, L. Bottou, Y. Bengio, and P. Haffner, Gradient-based learning applied to document recognition. In Proceedings of the IEEE, 86(11), pages 2278-2324, 1998

\bibitem{USPS} J. J. Hull, A database for handwritten text recognition research. In IEEE Transactions on Pattern Analysis and Machine Intelligence, 16(5), pages 550-554, May 1994

\bibitem{CoMNIST} Cyrillic oriented MNIST: A dataset of Latin and Cyrillic letter images. Website of Kaggle Inc., retrieved 22nd October 2018. https://www.kaggle.com/gregvial/comnist.











\end{thebibliography}

\end{footnotesize}


\end{document}